\documentclass[runningheads]{llncs}
\usepackage{graphicx}

\usepackage{tikz}
\usepackage{comment}
\usepackage{amsmath,amssymb} 
\usepackage{color}

\usepackage[accsupp]{axessibility}  

\usepackage[colorlinks,linkcolor=black,citecolor=black,urlcolor=black]{hyperref}

\usepackage{subcaption}
\usepackage{multirow}
\usepackage{hhline}
\usepackage{enumitem}
\usepackage[T1]{fontenc}
\usepackage{caption}

\newlength\savewidth\newcommand\shline{\noalign{\global\savewidth\arrayrulewidth\global\arrayrulewidth 1pt}\hline\noalign{\global\arrayrulewidth\savewidth}}

\begin{document}
\pagestyle{headings}
\mainmatter
\def\ECCVSubNumber{7756}  

\title{Dynamic Temporal Filtering in Video Models} 

\titlerunning{Dynamic Temporal Filtering in Video Models}
\author{Fuchen Long\inst{1}$^\star$\and
	Zhaofan Qiu\inst{1}\thanks{Fuchen Long and Zhaofan Qiu contributed equally to this work.} \and
	Yingwei Pan\inst{1} \and
	Ting Yao\inst{1} \and \\
	Chong-Wah Ngo\inst{2} \and
	Tao Mei\inst{1}}
\authorrunning{F. Long, Z. Qiu, Y. Pan, T. Yao, C.W. Ngo and T. Mei}
%
\institute{JD Explore Academy, Beijing, China \and Singapore Management University, Singapore\\
\email{\{longfc.ustc, zhaofanqiu, panyw.ustc, tingyao.ustc\}@gmail.com; cwngo@smu.edu.sg; tmei@jd.com}}
\maketitle

\begin{abstract}
Video temporal dynamics is conventionally modeled with 3D spatial-temporal kernel or its factorized version comprised of 2D spatial kernel and 1D temporal kernel. The modeling power, nevertheless, is limited by the fixed window size and static weights of a kernel along the temporal dimension. The pre-determined kernel size severely limits the temporal receptive fields and the fixed weights treat each spatial location across frames equally, resulting in sub-optimal solution for long-range temporal modeling in natural scenes. In this paper, we present a new recipe of temporal feature learning, namely Dynamic Temporal Filter (DTF), that novelly performs spatial-aware temporal modeling in frequency domain with large temporal receptive field. Specifically, DTF dynamically learns a specialized frequency filter for every spatial location to model its long-range temporal dynamics. Meanwhile, the temporal feature of each spatial location is also transformed into frequency feature spectrum via 1D Fast Fourier Transform (FFT). The spectrum is modulated by the learnt frequency filter, and then transformed back to temporal domain with inverse FFT. In addition, to facilitate the learning of frequency filter in DTF, we perform frame-wise aggregation to enhance the primary temporal feature with its temporal neighbors by inter-frame correlation. It is feasible to plug DTF block into ConvNets and Transformer, yielding DTF-Net and DTF-Transformer. Extensive experiments conducted on three datasets demonstrate the superiority of our proposals. More remarkably, DTF-Transformer achieves an accuracy of 83.5\% on Kinetics-400 dataset. Source code is available at \url{https://github.com/FuchenUSTC/DTF}.
\end{abstract}

\section{Introduction}
\begin{figure}[!tb]
	\centering\includegraphics[width=0.95\textwidth]{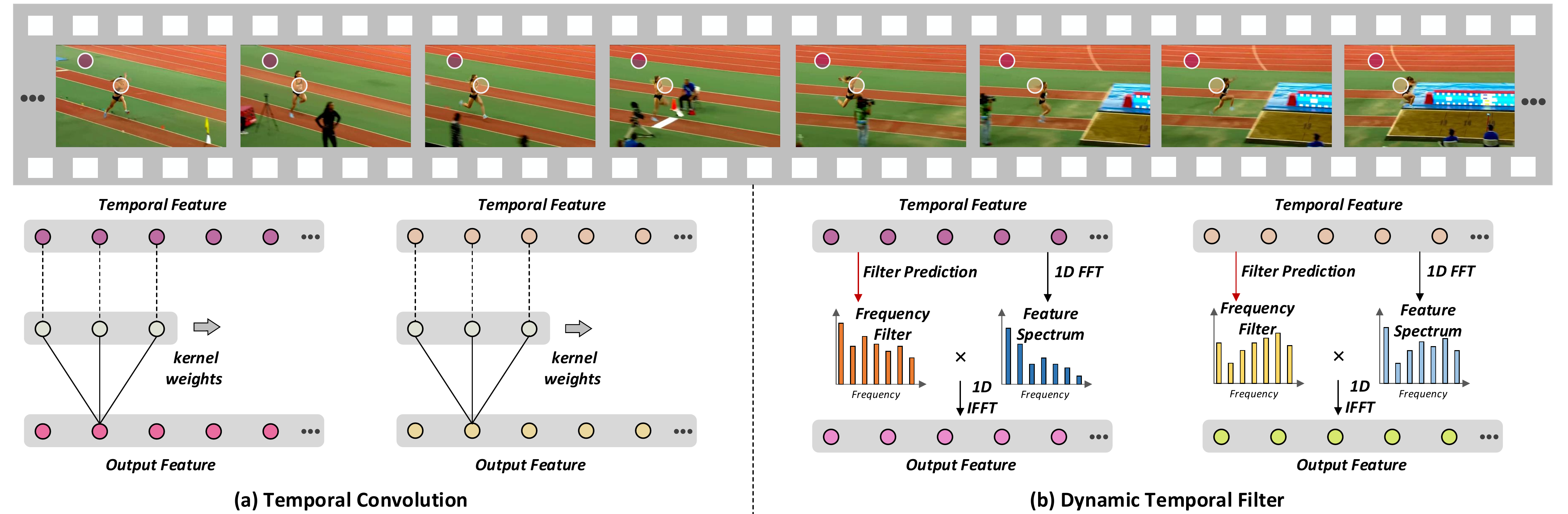}
    \vspace{-0.05in}
	\caption{Modeling the temporal evolution of two spatial regions (marked in pink and orange on the filmstrip) via (a) temporal convolution and (b) our DTF.}
	\label{fig1:1}
    \vspace{-0.25in}
\end{figure}

Video is an electronic carrier that records the evolution of moving persons or objects.
Modeling such evolution with time is essential to the understanding of motion patterns in videos.
The recent advances generally hinge on temporal convolution for temporal modeling in video models.
Furthermore, the common recipe is to integrate temporal convolution into space-time 3D convolution \cite{Ji:PAMI,Tran:ICCV15} or explicitly utilize temporal convolution to co-work with spatial convolution.
Figure \ref{fig1:1}(a) conceptually depicts the temporal modeling processes with 1D temporal convolution for two different spatial locations within the input video. The temporal convolution locally aggregates the features of the same spatial location in adjacent frames. The temporal receptive field is thus fixed and the corresponding kernel weights are the same across different spatial locations. This setting inevitably limits the temporal receptive field and ignores the inherent differences of spatial contexts at varied locations during temporal modeling.
Figure \ref{fig1:1} illustrates two spatial locations exhibiting different spatial contents: the pink dot refers to the track in the static background, while the orange dot shows a person moving rapidly across a constantly changing background. As the evolutions of spatial locations correspond to motion patterns specific to different movements, an optimal way of modeling is by having different kernels with varying size and weights that characterize their respective spatial context with sufficient temporal receptive field.
Having the same kernel filters over different spatial locations will hurt the mining of long-range temporal dependency.

In this paper, we propose to mitigate these issues by formulating the temporal feature learning in the frequency domain, pursuing a dynamic spatial-aware temporal modeling with an enlarged temporal receptive field. Specifically, we design a dynamic temporal filter (see Figure \ref{fig1:1}(b)) to characterize temporal evolution by learning frequency filter to adaptively modulate the spectrum of temporal features at different spatial locations. According to the convolution theorem \cite{SigSys}, the point-wise multiplication of spectrums in frequency domain of two signals is equivalent to the temporal convolution between them. As such, considering that the learnt specialized frequency filter operates over all the frequencies, this frequency filter can be interpreted as a temporal convolution with a larger kernel size. The design nicely enhances the mining of long-term temporal dependency in the frequency domain, without increasing computational/memory overhead. Meanwhile, in an effort to deal with different contexts of spatial locations, we dynamically learn a specialized frequency filter for each spatial location based on its temporal features across time. Furthermore, frame-wise aggregation is uniquely exploited to strengthen the primary temporal feature of each location by accumulating its temporal neighbors with inter-frame attention, thereby facilitating the learning of frequency filter.

By the frequency domain temporal modeling conditioned on the dynamic change of spatial contexts, we present a novel Dynamic Temporal Filter (DTF) block in video models. Technically, we regard the features sliced across frames at a fixed spatial location as the temporal feature. In order to enhance the primary temporal feature, we measure the temporal correlation between adjacent frames to estimate the motion clues, which are further utilized for aggregating temporal neighbors with inter-frame correlation. With this enhanced temporal feature, the specialized frequency filter can be more effectively learnt to capture context surrounding a spatial location.
At the same time, DTF converts the enhanced temporal feature of each spatial location into frequency feature spectrum via Fast Fourier Transform (FFT), which is further multiplied with the learnt frequency filter. Finally, inverse FFT is employed to reconstruct the temporal features from the modulated feature spectrum in frequency domain.

The DTF block can be viewed as a principled temporal modeling module, and is an alternative to standard 1D temporal convolution in the existing video backbones, such as CNN-based model or Transformer-based model, with favorable computation overhead. By directly inserting DTF block into the conventional 2D ResNet \cite{Kaiming:CVPR16} and Swin Transformer \cite{Swin-ViT}, we construct two kinds of new video backbones, i.e., DTF-Net and DTF-Transformer. Through extensive experiments over a series of action recognition benchmarks, we show that our DTF-Net and DTF-Transformer outperform the state-of-the-art video backbones.

\section{Related Work}
We group the recent temporal modeling techniques into two directions: hand-crafted based methods and deep model based methods, where the latter group can be further categorized into CNN-based and Transformer-based approaches.

\textbf{Hand-crafted Video Modeling}. The early works \cite{Klaser:BMVC08,Laptev:IJCV05,Laptev:CVPR08,Scovanner:MM07} construct hand-crafted video feature in two steps: detecting spatial-temporal interest points and formulating it by local descriptors. Trajectory is then adopted to convey motion cues. One is dense trajectory \cite{Wang:CVPR11} that samples local patch-wise features at various scales and tracks them through optical flow. Nevertheless, such features are not optimized, thereby hardly to be generalized across different tasks.

\textbf{CNN-based Video Modeling.} Early attempts for video CNN commonly apply 2D CNN for video input. Karpathy \emph{et al.} \cite{Sports1M} leverage spatial CNN to learn video representation by temporally stacking frame-level features.
Two stream model \cite{Simonyan:NIPS14} is adopted to employ 2D CNN over the inputs of visual frames and optical flow separately. Many variants \cite{Diba:CVPR17,Feichtenhofer:CVPR16,Wang:ECCV16,Yao:AAAI21} extend it in different aspects.
To address the long-range modeling issue ignored by two stream models, LSTM-based networks \cite{Yue-Hei:CVPR15,Srivastava:ICML15} are proposed to capture temporal dynamics in videos.
The above approaches only treat video as a sequence of frames, but leaving the pixel-level temporal evolution across consecutive frames unexploited.
The pioneering work of 3D CNN (C3D \cite{Tran:ICCV15}) is thus proposed to alleviate this issue. Furthermore, most subsequent research works \cite{Carreira:CVPR17,Feichtenhofer:CVPR20,Long:BCN,Qiu:ICCV17,Qiu:CVPR19,Tran:CVPR18,Xie:ECCV18,Zhao:NIPS18} found that disentangling spatial and temporal convolution leads to better performances against original 3D convolution and presents good generalization ability on localization tasks \cite{Long:CVPR19,Long:ECCV20,Long:TMM20}.
However, CNN-based methods still face the challenge of long-range modeling and fail to handle the inherent differences of spatial contexts.

\textbf{Transformer-based Video Modeling.} Inspired by the success of Vision Transformer (ViT) \cite{ViT} in image recognition, a series of Transformer-based backbones start to emerge.
Various variants \cite{TinT,Li:PAMI,Swin-ViT,GF-ViT,Touvron:DeiT,T2T-ViT} validated the power of self-attention for image feature learning. Similarly, the popularity of image Transformer leads to the investigation of the video Transformer architectures \cite{ViViT,Bertasius:ICML21,Fan:MVIT,Liu:V-Swin,Long:CVPR22}. TimeSformer \cite{Bertasius:ICML21} explores five different structures of space-time attention and suggests a factorized version for speed-accuracy tradeoff. MViT \cite{Fan:MVIT} further provides an alternative that formulates the video Transformer in a multi-scale manner. The pyramid features of MViT capture low-level visual information and high-level complex information.
Our DTF block is a temporal modeling primitive and can be readily pluggable to the 2D Transformer for video learning.

In short, our work belongs to the deep model based video modeling.
Unlike the traditional temporal convolution with a fixed kernel size that treats each spatial location equally, DTF performs convolution by dynamic spectrum filtering of each location in frequency domain with an enlarged receptive field.
Moreover, DTF block enhances the primary temporal features through frame-wise feature aggregation and provides additional motion clues for frequency filter prediction.

\section{Approach}
We introduce a new Dynamic Temporal Filter (DTF) for temporal modeling. Motivated by convolution theorem, DTF aims to convert temporal convolution to spectrum filtering in frequency domain.
Concretely, a novel temporal feature learning block, i.e, DTF block, is designed to perform such spectrum filtering for every spatial location in a video. The frequency filter is location-dependent for exploiting over-time contexts. By plugging DTF block into CNN and Transformer, we derive two video backbones, i.e., DTF-Net and DTF-Transformer.

\subsection{Preliminaries: Convolution Theorem}
\begin{figure}[!tb]
	\centering\includegraphics[width=0.96\textwidth]{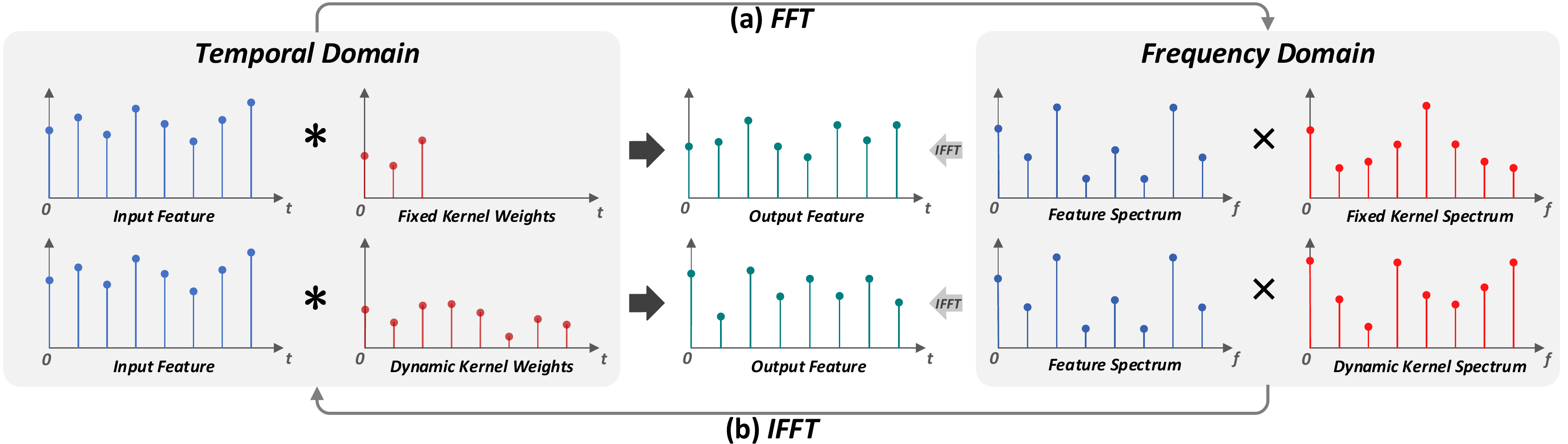}
	\caption{Illustration of (a) transformation from 1D convolution learning with a fixed kernel to the equivalent spectrum filtering in frequency domain via FFT (upper), and (b) transformation from spectrum filtering in frequency domain to the equivalent 1D convolution learning with a dynamic kernel via IFFT (lower). The temporal feature signals (and its' spectrum) and 1D convolution kernel (and its' spectrum) are represented as blue and red points, respectively.}
	\label{fig1:2}
    \vspace{-0.2in}
\end{figure}

To better understand the spirit of our DTF design, we first revisit the convolution theorem \cite{SigSys} in digital signal processing field. Formally, given a sequence of $T$ points feature signals ($f[t], 0\leq t\leq T-1$), its discrete spectrum $S[k]$ is calculated by Discrete Fourier Transform (DFT) as follows:
\begin{equation}\label{eq:3:1}
	S[k] = \sum_{t=0}^{T-1} f[t] e^{-j(2\pi/T)kt},~ 0 \leq k \leq T-1,
\end{equation}
where $j$ is the imaginary unit. Here DFT is a kind of one-to-one orthogonality decomposition. Furthermore, based on the DFT outputs, inverse DFT (IDFT) is able to reconstruct the input signals:
\begin{equation}\label{eq:3:2}
 f[t] = \frac{1}{T} \sum_{k=0}^{T-1} S[k] e^{j(2\pi/T)kt},~ 0 \leq t \leq T-1.
\end{equation}
Similarly, we achieve the spectrum $S_c[k]$ of the convolution kernel signal $f_c[t]$ via Fourier Transform.
The convolution theorem states that the Fourier Transform of a convolution of two signals is equivalent to the product of their Fourier Transformers. As shown in Figure \ref{fig1:2}, the output feature of convolution learning between the input feature and 1D kernel in temporal domain can be also learnt by the multiplication between their transformed spectrum through IDFT:
\begin{equation}\label{eq:3:3}
	f[t] \ast f_c[t] = IDFT(S[k] \times S_c[k]),
\end{equation}
where $\ast$ and $\times$ denotes convolution and element-wise multiplication, respectively.

Given a real convolution kernel with fixed size, the corresponding kernel spectrum is conjugate symmetric (see the top-right part in Figure \ref{fig1:2}). This implies that only half of the spectrum points ($M=\lfloor T/2 \rfloor+1$) are capable of covering all information of frequency property. In other words, there exists information redundancy in kernel spectrum. To address this issue, we propose to learn a dynamic filter in frequency domain to modulate the feature spectrum. Specifically, as depicted in Figure \ref{fig1:2} (lower part), when multiplying the frequency feature spectrum with a frequency filter (i.e., kernel spectrum with varied contents in all frequencies), this process is equivalent to the convolution learning between input feature and dynamic kernel with an enlarged temporal receptive field.

For implementation of DFT, the Fast Fourier Transform (FFT) is commonly employed for engineering. The corresponding inverse DFT is thus implemented as inverse Fast Fourier Transform (IFFT). Thus, we choose FFT and IFFT as the basic transformation in the architecture of our Dynamic Temporal Filter.

\subsection{Dynamic Temporal Filter (DTF)} \label{sec:3.2}
Most existing temporal modeling approaches employ 1D temporal convolution to perform pixel-level aggregation across frames. Nevertheless, the pre-determined kernel size of the 1D temporal convolution severely limits the mining of long-range dependency. Meanwhile, typical temporal convolution treats each spatial position equally, and thus ignores the inherent differences of spatial contexts at varied locations.
Inspired by convolution theorem, we novelly formulate the temporal modeling in frequency domain. A new Dynamic Temporal Filter (DTF) mechanism is thus designed to learn a specialized frequency filter based on the context of each spatial location for modulating the frequency feature spectrum.

\begin{figure}[!tb]
	\centering\includegraphics[width=0.95\textwidth]{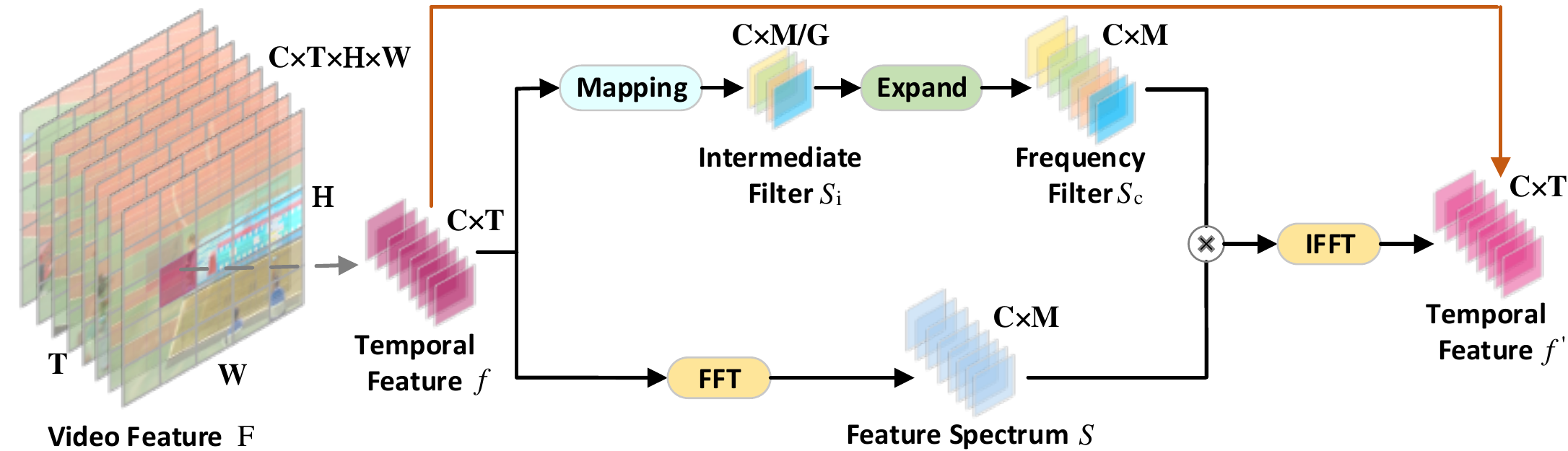}
    \vspace{-0.1in}
	\caption{Illustration of Dynamic Temporal Filter (DTF) mechanism.}
	\label{fig1:3}
    \vspace{-0.2in}
\end{figure}

Here we introduce the detailed formulation of DTF mechanism (see Figure \ref{fig1:3}). Let $F$ be the input 3D feature map with the size of $C\times T\times H\times W$, where $C$, $H\times W$, and $T$ denotes the channels size, spatial size and temporal length, respectively. For each spatial location, we take the feature cube across time at that location in $F$ as the temporal feature $f \in \mathbb{R}^{C\times T}$. For each channel in $f$, we apply Fast Fourier Transform (FFT) along the temporal dimension to obtain the whole feature spectrum $S \in \mathbb{C}^{C\times M}$. Please note that the point number of the spectrum of a real signal is $M=\lfloor T/2 \rfloor+1$ and it is in the field of the complex numbers. Meanwhile, a specialized frequency filter $S_c \in \mathbb{C}^{C\times M}$ is learnt conditioned on the temporal feature $f$. It is natural to implement the estimator of frequency filter as a fully connected layer. However, directly predicting the frequency filter through linear mapping requires heavy memory overhead ($C^2 \times T\times M$ parameters). Hence we take the inspiration from group convolution, and significantly reduce the parameters of estimator by sharing some temporal filters across channels. Most specifically, we first project $f$ into the intermediate filter $S_i \in \mathbb{C}^{C\times M/G}$, where the number of channels is decreased by a factor of $G$. After that, $S_i$ is expanded along the channel dimension to achieve the complete frequency filter $S_c$. Next, we modulate the frequency feature spectrum $S$ by multiplying it with the learnt frequency filter $S_c$, leading to the modulated feature spectrum $S'$:
\begin{equation}\label{eq:3:4}
	S' = S \times S_c.
\end{equation}
Then, we adopt the inverse FFT to transform the modulated spectrum $S'$ into the video feature $f'$ in the temporal domain:
\begin{equation}\label{eq:3:5}
	f' = IFFT(S').
\end{equation}
Finally, the output temporal feature $f_o$ of DTF is achieved by fusing the original temporal feature $f$ and the modulated temporal feature $f'$ as $f_o = f + f'$.

Accordingly, DTF mechanism triggers temporal modeling in the frequency domain by modulating the spectrum of temporal features with the learnt frequency filters.
Compared to traditional 1D temporal convolution with fixed kernel size, the enlarged temporal receptive field derived from a learnt frequency filter in DTF strengthens the long-range dependency modeling. Moreover, unlike using the same kernel weights in 1D temporal convolution for all spatial locations, our DTF learns specialized frequency filter based on the different context of each spatial location, pursuing a dynamic spatial-aware temporal modeling.

\subsection{DTF Block}
Recall that our DTF mechanism characterizes the temporal evolution of each same spatial location across time in frequency domain by learning spatial-aware frequency filter. However, this way inevitably ignores the rich contextual information between each spatial location and its temporal neighbors in adjacent frames for temporal modeling. To alleviate this issue, we devise a DTF Block that capitalizes on a self-attention based frame-wise aggregation (FA) approach before DTF mechanism to enhance temporal features, which also provides additional motion clues to boost the learning of frequency filters.

\begin{figure}[!tb]
	\centering\includegraphics[width=0.95\textwidth]{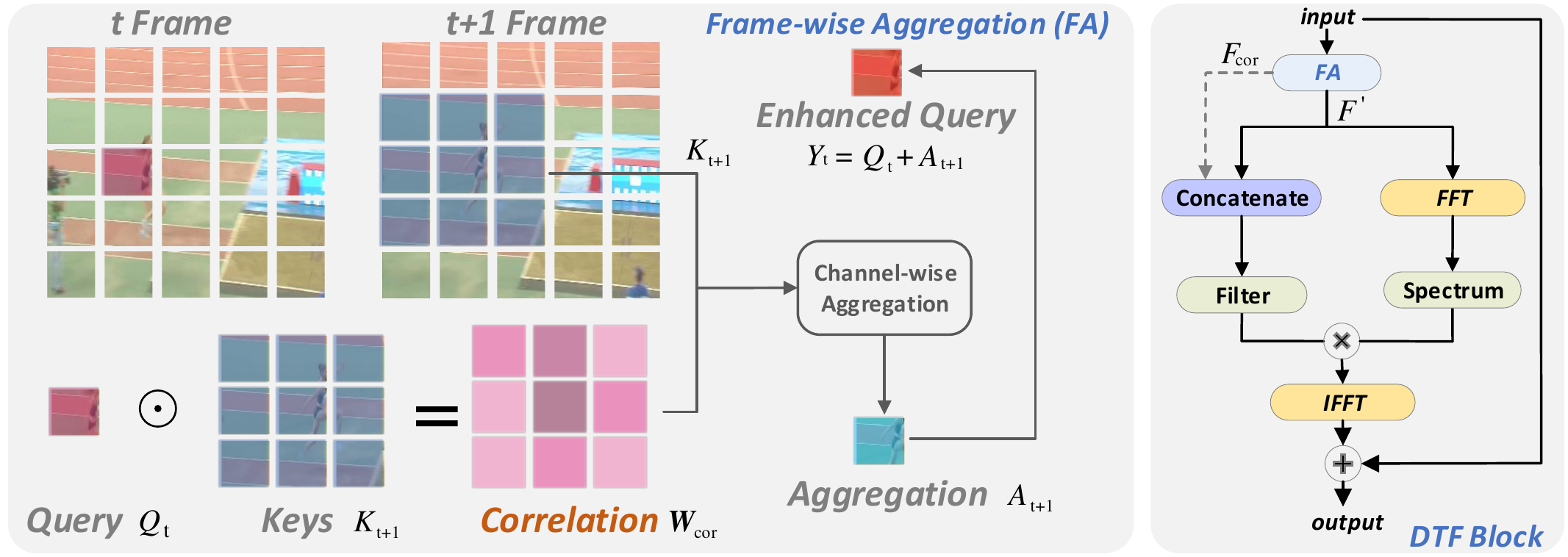}
	\caption{The architectures of Frame-wise Aggregation (FA) and our DTF block.}
	\label{fig1:4}
    \vspace{-0.2in}
\end{figure}

Technically, inspired by self-attention learning \cite{Vaswani:NIPS17,Wang:CVPR18}, we first strengthen primary temporal feature of each spatial location by exploring inter-frame interaction and aggregating its temporal neighbors in adjacent frames. Figure \ref{fig1:4} (left) details the process of the frame-wise aggregation in DTF block. Specifically, given the input 3D feature map $F \in \mathbb{R}^{C\times T\times H\times W}$, we take the feature at spatial location ($x,y$) of $t$-th frame as the query $Q_t \in \mathbb{R}^C$. For $Q_t$, all its temporal neighbors in ($t$+1)-th frame within the local region ($k\times k$ grid) centered at ($x,y$) are set as keys $K_{t+1} \in \mathbb{R}^{C\times \{k\times k\}}$. After that, we achieve the inter-frame correlation matrix $\mathbf{W}_{cor} \in\mathbb{R}^{1\times \{k\times k\}}$ via self-attention:
\begin{equation}\label{eq:3:7}
	\mathbf{W}_{cor} = Q_t \odot K_{t+1},
\end{equation}
where $\odot$ is the matrix multiplication that measures the similarity between query $Q_t$ and its' temporal neighbors $K_{t+1}$ within the region of $k\times k$ grid. We further utilize the inter-frame correlation matrix as the attention weights to aggregate temporal neighbors $K_{t+1}$ within the ($t$+1)-th frames as follows:
\begin{equation}\label{eq:3:8}
	{A}_{t+1} = \mathbf{W}_{cor} \odot [K_{t+1}]^{Tr},
\end{equation}
where $A_{t+1}$ is the aggregated temporal feature and $[\cdot]^{Tr}$ denotes the operation of matrix transposition.
The aggregated feature is further employed to strengthen the query feature, and the enhanced query feature $Y_t$ is thus measured as:
\begin{equation}\label{eq:3:9}
	{Y}_t = Q_t + A_{t+1}.
\end{equation}

We operate frame-wise aggregation between every pair of consecutive frames, yielding the enhanced video representation $F' \in \mathbb{R}^{C\times T\times H\times W}$. Next, DTF mechanism takes the enhanced temporal feature $F'$ as inputs, and transforms it into feature spectrum via FFT for frequency modulation. Considering that the inter-frame correlation $\mathbf{W}_{cor}$ reflects the pixel-level displacement information, we exploit it as additional guidance to strengthen the learning of frequency filter.
In particular, as shown in Figure \ref{fig1:4} (right), DTF block directly squeezes the learnt correlation weights $\mathbf{W}_{cor}$ of all the temporal neighbors in FA as the correlation feature $F_{cor} \in \mathbb{R}^{k^2\times T\times H\times W}$. Then, the enhanced temporal feature $F'$ is concatenated with the correlation feature $F_{cor}$ for learning the specialized temporal filter of each spatial location. In this way, DTF block additionally mines the motion clues from the correlation feature in FA to facilitate frequency filter prediction.

\subsection{Video Backbones with DTF Block}
\begin{figure}[!tb]
	\centering\includegraphics[width=0.90\textwidth]{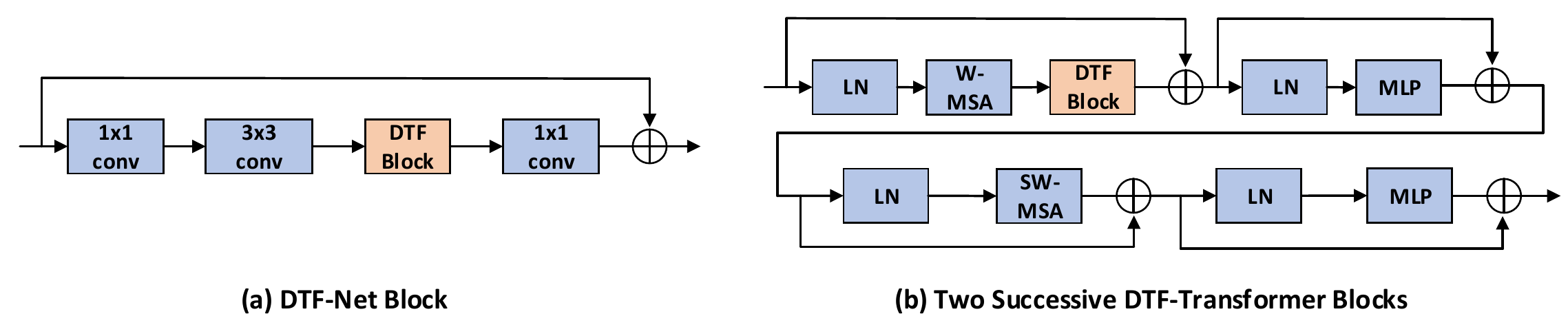}
	\caption{Basic blocks in (a) DTF-Net and (b) DTF-Transformer.}
	\label{fig1:5}
    \vspace{-0.2in}
\end{figure}

Our DTF block is readily pluggable to existing 2D CNN or Vision Transformer to upgrade the vision backbones for video temporal modeling. Here we present how to insert DTF block into ResNet \cite{Kaiming:CVPR16} and Swin Transformer \cite{Swin-ViT}. Figure \ref{fig1:5} depicts two different constructions of DTF block in building block in ResNet/Swin Transformer, namely DTF-Net and DTF-Transformer, respectively.

\textbf{DTF-Net.} Most of video architecture advances \cite{Carreira:CVPR17,Qiu:ICML21,Tran:CVPR18,Xie:ECCV18} typically factorize the conventional 3D convolution into 2D spatial convolution and 1D temporal convolution, and the 1D temporal convolution is commonly integrated after the spatial convolutional layers of 2D CNN for temporal modeling across frames. We follow this recipe and construct the DTF-Net by inserting DTF-Block after the $3\times3$ convolution within each basic residual building block in ResNet \cite{Kaiming:CVPR16}. Based on the output feature of the final residual building block, the global pooling is employed to achieve the clip-level feature for video representation learning.

\textbf{DTF-Transformer.} Recently, the Transformer-style architectures with self-attention \cite{ViT,Swin-ViT} have emerged as powerful backbones in compute vision field. Inspired by this, we further integrate DTF block into the Swin-Transformer \cite{Swin-ViT} to build the Transformer-style video backbone, named as DTF-Transformer. Specifically, for every two successive Swin Transformer blocks in Swin Transformer, we directly plug the DTF block after the multi-head self-attention module with regular window (W-MSA), leading to the two successive DTF-Transformer building block. Here we reshape the output patch sequence of W-MSA module into the sequence of frame feature map with the normal size ($C\times T\times H\times W$), and then feed it into the DTF block.
Global pooling is utilized to obtain clip-level feature.

\section{Experiments}
\subsection{Datasets and Implementation Details}
\textbf{Datasets.} We empirically evaluate the effectiveness of our proposed video backbones (DTF-Net and DTF-Transformer) on three datasets, i.e., \textbf{Kinetics-400} \cite{Carreira:CVPR17}, \textbf{Something-Something V1 and V2} \cite{Goyal:SS}.
The Kinetics-400 dataset is composed of 300K videos derived from 400 action categories. Each video is 10-seconds short clip cropped from the raw YouTube video.
We split all the 300K videos into 240K, 20K, 40K for training, validation and testing, respectively.
Something-Something V1 and V2 datasets include about 108K and 221K videos over 174 action categories.
For Something-Something V1 and V2, there are 86K/11.5K/11K and 169K/25K/27K videos in the training/validation/testing set, respectively.

\textbf{Network Training.} We implement our proposal on PyTorch framework. The mini-batch Stochastic Gradient Descent (SGD) algorithm with cosine learning rate \cite{Loshchilov:ICLR17} is utilized for network optimization.
The resolution of each frame is fixed as $224\times 224$, which is randomly cropped from the video clip resized with the short size in [$256,340$].
We set the input clip length within the range from $16$ to $64$.
Each clip is randomly flipped along horizontal direction for data augmentation, except for Something-Something V1 and V2 in view of the direction-related classes.
We set the size of the local region $k$ and the factor $G$ in DTF block as $3$ and $16$.
The base learning rate is $0.04$ for DTF-Net and $0.01$ for DTF-Transformer, respectively. The dropout ratio is fixed as $0.5$.
The maximum training epoch number is $128$/$64$ for Kinetics-40/Something-Something datasets.
The mini-batch size and the weight decay parameter is $256$ and $0.0001$.

\textbf{Network Inference.} Two kinds of inference strategies are adopted to evaluate DTF-Net and DTF-Transformer. For DTF-Net, we follow the 3-crop strategy \cite{Christoph:ICCV19} to crop three $256\times 256$ regions from each clip at inference. The video-level prediction is calculated by averaging all scores from 10 uniform sampled clips. For DTF-Transformer, we directly measure the video-level score based on the 4 uniform sampled clips. The 3-crop strategy is also adopted for score fusion.

\subsection{Ablation Study on DTF Block}
Here we perform ablation studies to examine each technical choice in DTF block of DTF-Net. Note that DTF-Net is constructed based on the ResNet-50, and we report the top-1 and top-5 accuracy on the validation set of Kinetics-400.

\begin{table*}[!tb]
	\centering
	\caption{Ablation study on DTF block in DTF-Net with 16-frame inputs on Kinetics-400 dataset. Top-1 and Top-5 accuracy (\%), and the computational cost (estimated by GFLOPs) for forwarding one clip at inference are reported.}
	\subcaptionbox{
		\label{tab1:dtf}
		\textbf{Dynamic Temporal Filter.} Comparisons among different variants of DTF. All runs are built by plugging each block into $res_5$ stage of ResNet-50.}{
		\scalebox{0.88}[0.78]{
			\begin{tabular}{l|c|c c}
				\shline
				Model                         & GFLOPs &  Top-1  & Top-5  \\ \shline
				2D-ResNet                     &  23  &  72.0   & 90.3   \\
				\hline
				DTF$_{1d}$                    &  25  &  73.2   & 90.7   \\
				DTF$_{1d+}$                   &  28  &  74.2   & 91.6   \\
				DTF$_{F}$					  &	 23  &  75.0   & 92.2   \\
				\hhline{*{4}{-}}
				DTF                           &  24  &  75.7   & 92.9  \\ \shline
			\end{tabular}
	}}
	\subcaptionbox{
		\label{tab1:agg}
		\textbf{Frame-wise Aggregation.} Effect investigation of frame-wise aggregation and the correlation feature in DTF block. All runs are constructed by plugging each block into $res_5$ stage of ResNet-50.}{
		\scalebox{0.88}[0.78]{
			\begin{tabular}{c c|c|c c}
				\shline
				\multicolumn{2}{c|}{Model} & \multicolumn{1}{|c|}{\multirow{2}{*}{\text{GFLOPs}}} &  \multicolumn{1}{|c}{\multirow{2}{*}{\text{Top-1}}}  &\multicolumn{1}{c}{\multirow{2}{*}{\text{Top-5}}}  \\ \hhline{*{2}{-}}
				Aggregation & Correlation & \multicolumn{1}{|c|}{} & \multicolumn{1}{|c}{} & \multicolumn{1}{c}{} \\
				\shline
				\multicolumn{2}{c|}{DTF-baseline}    &  23  &  74.9   & 92.3   \\
				\hline
				\checkmark   &  		  &  24   &  75.4   & 92.6   \\
				& \checkmark 			  &  24   &  75.2   & 92.5   \\
				\checkmark   & \checkmark &  24   &  75.7   & 92.9   \\ \shline
			\end{tabular}
	}}
	\subcaptionbox{
		\label{tab1:stage}
		\textbf{Location of DTF in DTF-Net.} Effect of plugging DTF blocks into different stages of ResNet-50.}{
		\scalebox{0.80}[0.70]{
			\begin{tabular}{c c c c|c|c c}
				\shline
				\multicolumn{4}{c|}{Stage} & \multicolumn{1}{|c|}{\multirow{2}{*}{\text{GFLOPs}}} &  \multicolumn{1}{|c}{\multirow{2}{*}{\text{Top-1}}}  &\multicolumn{1}{c}{\multirow{2}{*}{\text{Top-5}}}  \\ \hhline{*{4}{-}}
				$res_2$ & $res_3$ & $res_4$ & $res_5$ & \multicolumn{1}{|c|}{} & \multicolumn{1}{|c}{} & \multicolumn{1}{c}{} \\
				\shline
				\multicolumn{4}{c|}{2D-ResNet}   &  23   &  72.0   & 90.3   \\
				\hline
				&            &            & \checkmark      &  24   &  75.7   & 92.9   \\
				&            & \checkmark & \checkmark      &  24   &  76.5   & 93.0   \\
				& \checkmark & \checkmark & \checkmark      &  25   &  77.1   & 93.1  \\
				\checkmark & \checkmark & \checkmark & \checkmark &  25   &  \textbf{77.7}   & \textbf{93.2}   \\ \shline
			\end{tabular}
	}}
	\subcaptionbox{
		\label{tab1:temporal}
		\textbf{Temporal Modeling.} Comparisons among different temporal modeling methods based on ResNet-50 backbone.}{
		\scalebox{0.80}[0.70]{
			\begin{tabular}{l|c|c c}
				\shline
				Temporal Modeling                       & GFLOPs&  Top-1  & Top-5  \\ \shline
				2D-ResNet                               &  23  &  72.0   & 90.3   \\
				\hline
				Temporal Conv \cite{Tran:CVPR18}        &  33  &  74.6   & 91.5   \\
				Temporal Shift \cite{JiLin:ICCV19}      &  23  &  74.8   & 91.5 \\
				Correlation \cite{Wang:CVPR20}          &  23  &  75.1   & 91.6   \\
				Temporal Difference \cite{Wang:CVPR21}  &  36  &  76.6   & 92.8   \\
				\hhline{*{4}{-}}
				DTF                                     &  25   &  \textbf{77.7}   &  \textbf{93.2}  \\ \shline
			\end{tabular}
	}}  	
	\label{tab:ablations}
\end{table*}

\textbf{Dynamic Temporal Filter.}
We first evaluate how each design in our DTF block influences the overall performance of DTF-Net. Table \ref{tab1:dtf} details the performance comparisons among different variants of DTF block. Note that all runs here are implemented by inserting DTF variants into the basic residual blocks at $res_5$ stage of ResNet-50.
The run of \textbf{2D-ResNet} is regarded as a basic 2D bottleneck residual block and there is no temporal modeling.
By integrating the basic block with the conventional temporal 1D convolution \cite{Tran:CVPR18}, \textbf{DTF}$_{1d}$ obtains better performances, which demonstrate the advantage of temporally pixel-wise feature aggregation for motion modeling. Nevertheless, such operation employs the fixed weights over the feature cube of each spatial location. Instead, \textbf{DTF}$_{1d+}$ learns 1D dynamic convolution for each location (i.e., the 1D variant of dynamic convolution \cite{Chen:CVPR20}), and outperforms \textbf{DTF}$_{1d}$. The results basically indicate the merit of dynamic kernel learning, but this block brings a clear overhead in computation cost. Instead of temporal modeling in temporal domain, \textbf{DTF}$_F$ performs temporal modeling in frequency domain by modulating feature spectrum with a fixed frequency filter for each location. Benefiting from the equivalent enlarged temporal receptive field, \textbf{DTF}$_F$ further boosts up the performances. \textbf{DTF} additionally triggers the dynamic spatial-aware temporal modeling of each spatial location with specialized frequency filter, thereby leading to a performance gain by 0.7\% in top-1 accuracy with a slight computation overhead.

\textbf{Frame-wise Aggregation.}
Next, we investigate the effectiveness of the frame-wise aggregation and correlation feature in DTF block. Table \ref{tab1:agg} summarizes the performances across different variants of DTF block. DTF-baseline is the degraded version of DTF block without using frame-wise aggregation before FFT, which has already achieved 74.9\% top-1 accuracy on Kinetics-400. Next, by equipping DTF-baseline with frame-wise aggregation, the performance is further improved to 75.4\%. When solely exploring the correlation weights as the additional motion cues for frequency filter learning, the performance improvements against DTF-baseline are also attained. Furthermore, by simultaneously enhancing temporal feature via frame-wise aggregation and boosting filter learning with correlation weights, DTF block achieves the highest performances.

\textbf{Location of DTF block in DTF-Net.}
To examine the relationship between performance and the location of our DTF block in DTF-Net, we gradually plug DTF blocks into the stages in ResNet-50 backbone, and compare the performances.
The performance trend shown in Table \ref{tab1:stage} indicates that the 2D-ResNet benefits more by inserting DTF blocks into more stages and the increase of the computation cost is very slight.
Taking a closer look at the top-1 accuracy of different locations of DTF block, the injection of DTF blocks into the only one stage ($res_5$) already leads to a large improvement of 3.7\% against 2D-ResNet, which clearly validates the temporal modeling ability of DTF.
By further integrating all the four stages in ResNet-50 with DTF blocks, DTF-Net achieves the best performances, without requiring heavy computation overhead.

\textbf{Temporal Modeling.}
We next make the comparison between DTF and other existing temporal modeling techniques.
The performances of integrating the ResNet-50 backbone with different temporal modeling approaches are listed in Table \ref{tab1:temporal}.
Overall, our DTF leads to higher top-1 accuracy against other temporal modeling models with similar or even less computation cost.
The results basically demonstrate the advantage of exploring dynamic spatial-aware temporal modeling in frequency domain.
Specifically, by additionally modeling temporal dynamics via temporal convolution, Temporal Conv \cite{Tran:CVPR18} is superior to 2D-ResNet.
Correlation \cite{Wang:CVPR20} explicitly captures motion displacement across frames, and outperforms Temporal Conv.
By capturing long-range motion patterns through RGB/feature differences, Temporal Difference \cite{Wang:CVPR20} shows better performances than Correlation.
However, the performances of Temporal Difference are still below that of DTF block which dynamically modulates frequency feature spectrum with learnt frequency filter for temporal modeling.

\subsection{Evaluation on Long-Range Temporal Modeling}
\begin{figure}[!tb]
	\centering
	\begin{minipage}[t]{.48\linewidth}\centering
		\includegraphics[width=0.80\textwidth]{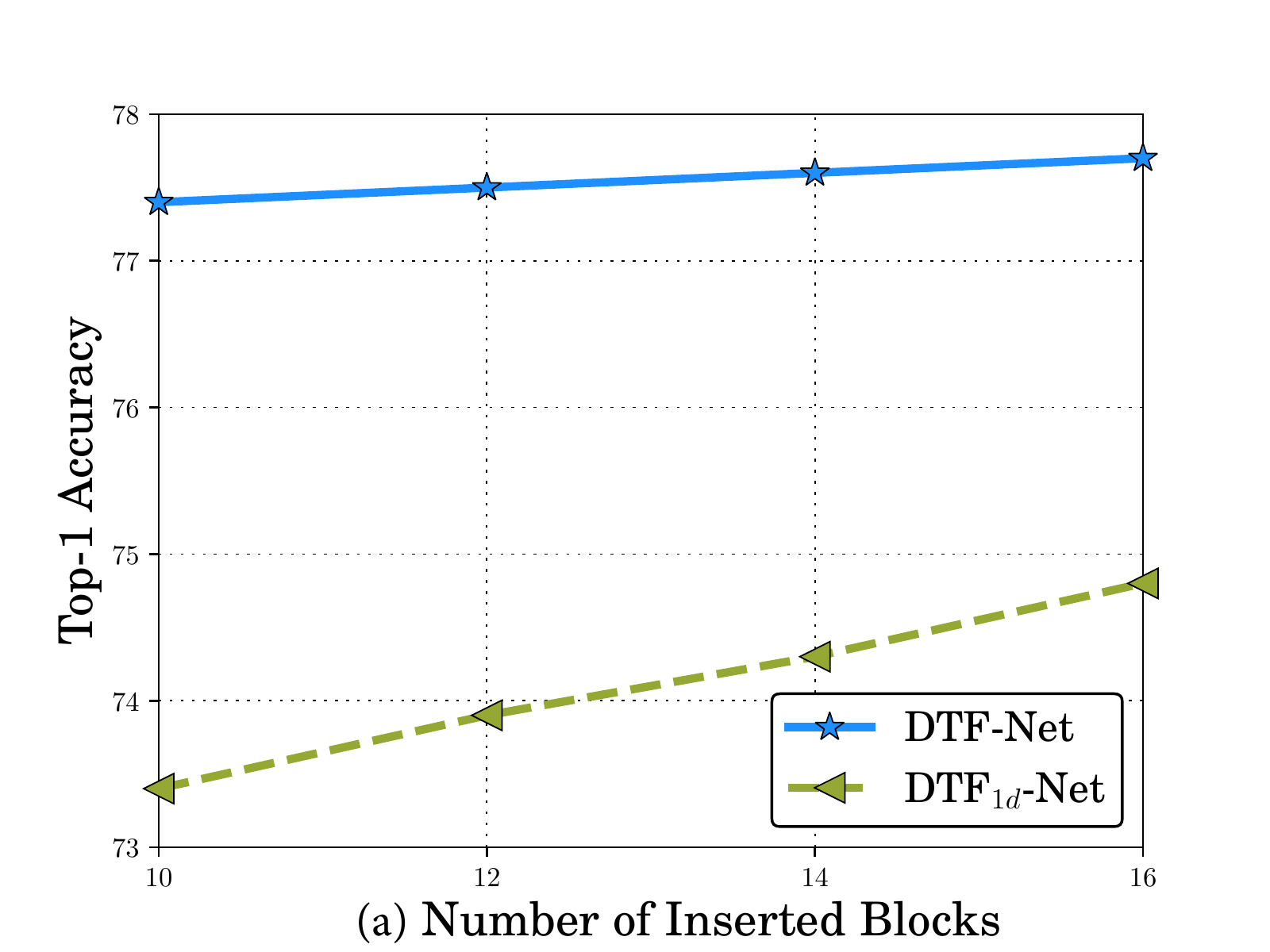}
	\end{minipage}
	\begin{minipage}[t]{.48\linewidth}\centering
		\includegraphics[width=0.80\textwidth]{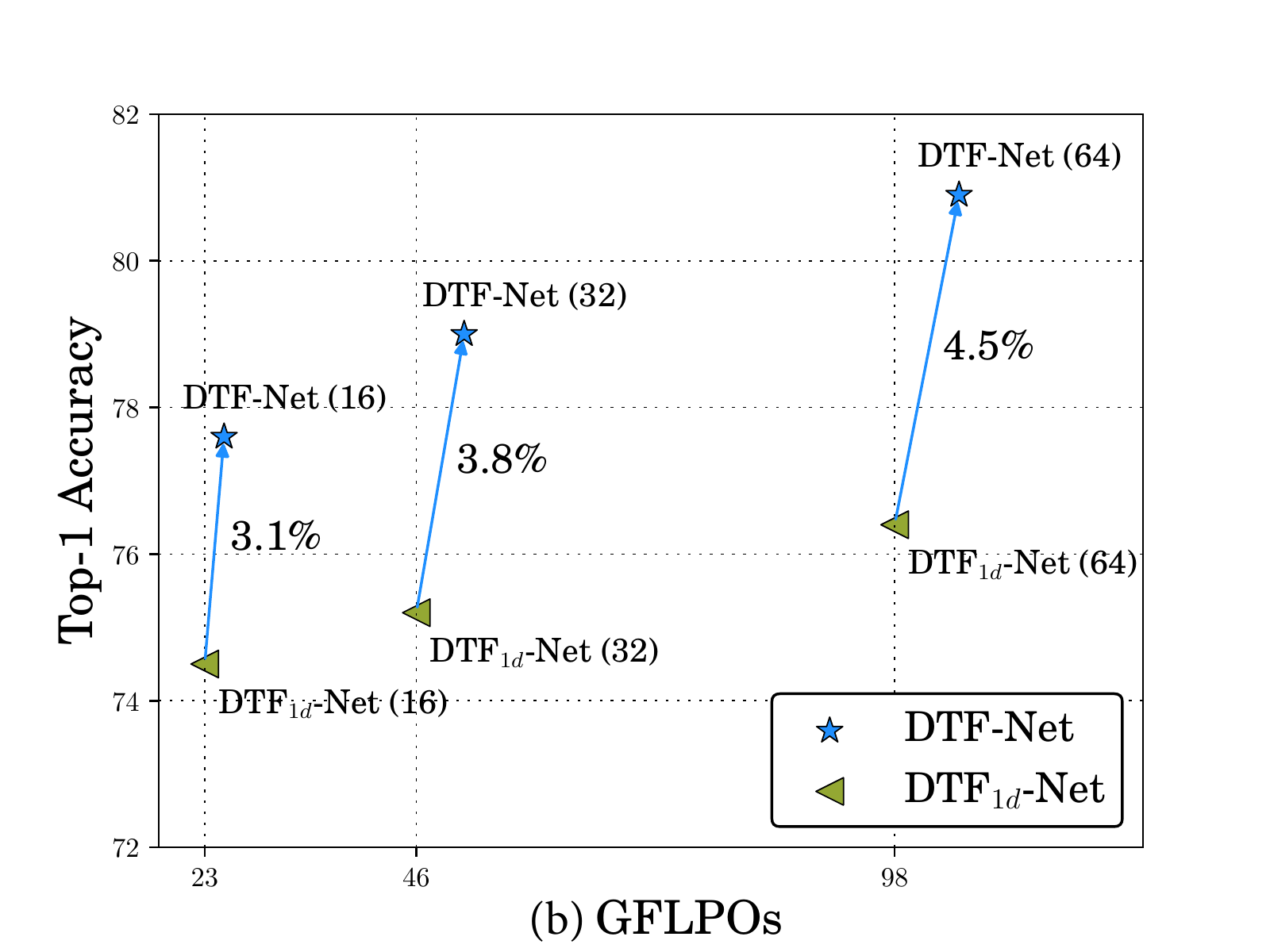}
	\end{minipage}
    \vspace{-0.12in}
	\caption{Performance comparisons between DTF-Net and DTF$_{1d}$-Net by using (a) different number of inserted blocks (with 16-frame input) and (b) different input clip length on Kinetics-400 (backbone: ResNet-50).}
	\label{fig1:6}
    \vspace{-0.13in}
\end{figure}

The commonly adopted temporal convolution in existing video backbones is characterized with the fixed kernel size and limited temporal receptive field. They often stack multiple temporal modeling blocks to expand the temporal receptive field for long-range temporal modeling. Instead, our DTF novelly formulates temporal modeling in frequency domain with enlarged temporal receptive field.
Therefore, even with a small number of inserted DTF blocks, DTF-Net should be still capable of capturing long-range dependencies.
Moreover, DTF should benefit more from the longer input clip length through the dynamic temporal modeling.
To validate these claims, we empirically compare the performances between DTF$_{1d}$-Net and DTF-Net on Kinetics-400 when capitalizing on different number of inserted blocks and input clip length in Figure \ref{fig1:6}. Note that DTF$_{1d}$-Net is a degradation of DTF-Net by employing conventional temporal 1D convolution in each basic residual block for temporal modeling.
As shown in this figure, DTF-Net consistently outperforms DTF$_{1d}$-Net across different number of blocks and different number of input frames.
More specifically, in Figure \ref{fig1:6}(a), the accuracy of DTF$_{1d}$-Net decreases more sharply than that of DTF-Net when reducing the number of inserted blocks.
Meanwhile, in Figure \ref{fig1:6}(b), the performance gap between DTF$_{1d}$-Net and DTF-Net is increased when feeding into more frames.
Both of the results confirm the merit of exploring temporal modeling in frequency domain to capture long-range dependency.

\begin{table}[!tb]
	\setlength{\belowcaptionskip}{-1pt}
	\caption{Performance comparisons with state-of-the-art video backbones on Kinetics-400. The input clip length of DTF-Net is shown inside the bracket.}
	\begin{minipage}[!tb]{.5\linewidth}
		\centering
		\scalebox{0.64}[0.58]{
			\begin{tabular}{{l|l|c|c|c}}
				\shline
				\multicolumn{1}{c|}{\textbf{Approach}} & \textbf{Backbone} & \textbf{GFLOPs$\times$views} & \textbf{Top-1} & \textbf{Top-5} \\ \shline
				\multicolumn{5}{l}{\textbf{Convolutional Networks}} \\ \shline
				\text{I3D \cite{Carreira:CVPR17}}                       &  Inception & 108$\times$N/A  & 72.1  & 90.3 \\
                \text{TSN \cite{Wang:ECCV16}}                           &  Inception & 80$\times$10    & 72.5  & 90.2  \\
				\text{MF-Net \cite{Chen:ECCV18}}                        &  R34       & 11$\times$50    & 72.8  & 90.4 \\
				\text{R(2+1)D \cite{Tran:CVPR18}}                       &  R34       & 152$\times$10   & 74.3  & 91.4  \\
				\text{S3D \cite{Xie:ECCV18}}                            &  Inception & 71$\times$30    & 74.7  & 93.4  \\
				\text{TSM \cite{JiLin:ICCV19}}                          &  R50       & 33$\times$30    & 74.1  & 91.2   \\
				\text{TEINet \cite{Liu:AAAI20}}                         &  R50       & 33$\times$30    & 74.9  & 91.8   \\
				\text{TEA \cite{Yan:CVPR20}}                            &  R50       & 33$\times$30    & 75.0  & 91.8   \\
				\text{SlowFast \cite{Christoph:ICCV19}}                 &  R50+R50   & 36$\times$30    & 75.6  & 92.1  \\
				\text{NL I3D \cite{Wang:CVPR18}}                        &  R50       & 282$\times$30   & 76.5  & 92.6   \\
				\text{SmallBig \cite{Li:CVPR20}}                        &  R50       & 57$\times$30    & 76.3  & 92.5  \\
				\text{CorrNet \cite{Wang:CVPR20}}                       &  R50       & 115$\times$10   & 77.2  & -     \\
				\text{TDN \cite{Wang:CVPR21}}                           &  R50       & 72$\times$30    & 77.5  & 93.2  \\ \hline
				\text{DTF-Net (16)}							            &  R50	     & 25$\times$30    & {77.7}  & {93.2}    \\
				\text{DTF-Net (32)}							            &  R50       & 51$\times$30    & {78.9}  & {93.8}    \\
				\text{DTF-Net (64)}							            &  R50	     &111$\times$30    & \textbf{80.9}  & \textbf{94.6}    \\ \shline
			\end{tabular}
		}
	\end{minipage}
	\begin{minipage}[!tb]{.5\linewidth}
		\centering
		\scalebox{0.65}[0.61]{
			\begin{tabular}{{l|l|c|c|c}}
				\shline
				\multicolumn{1}{c|}{\textbf{Approach}} & \textbf{Backbone} & \textbf{GFLOPs$\times$views} & \textbf{Top-1} & \textbf{Top-5} \\ \shline
				\multicolumn{5}{l}{\textbf{Convolutional Networks}} \\ \shline
				\text{ip-CSN \cite{Tran:ICCV19}}                        &  R101      & 83$\times$30    & 76.7    & 92.3   \\
				\text{SmallBig \cite{Li:CVPR20}}                        &  R101      & 418$\times$12   & 77.4    & 93.3   \\
				\text{NL I3D \cite{Wang:CVPR18}}                        &  R101      & 359$\times$30   & 77.7    & 93.3    \\
				\text{TDN \cite{Wang:CVPR21}}                           &  R101      & 132$\times$30   & 78.5    & 93.9   \\
				\text{CorrNet \cite{Wang:CVPR20}}                       &  R101      & 224$\times$30   & 79.2    &  -     \\
				\text{SlowFast \cite{Christoph:ICCV19}}                 &  R101+R101 & 234$\times$30   & 79.8    & 93.9   \\
				\hline
				\text{DTF-Net (16)}                                     &  R101      &  38$\times$30  & {78.9}  & {94.1}   \\
				\text{DTF-Net (32)}							            &  R101      &  76$\times$30  & {80.1}  & {94.3}   \\
				\text{DTF-Net (64)}							            &  R101      &  152$\times$30 & \textbf{81.8}  & \textbf{94.9}   \\
				\shline		
				\multicolumn{5}{l}{\textbf{Vision Transformer}} \\ \shline
				\text{TimeSformer \cite{Bertasius:ICML21}}              &  ViT-B     & 2,380$\times$3   & 80.7  & 94.7 \\
				\text{ViViT \cite{ViViT}}                               &  ViT-L     & 3,992$\times$12  & 81.3  & 94.7 \\
				\text{MViT \cite{Fan:MVIT}}                             &  MViT-B    & 455$\times$9     & 81.2  & 95.1 \\
				\text{Video-Swin \cite{Liu:V-Swin}}                     &  Swin-B    & 282$\times$12    & 82.7  & 95.5 \\
				\hline
				\text{DTF-Transformer}                                  &  Swin-B    & 266$\times$12    & \textbf{83.5}  & \textbf{95.9} \\
				\shline	
			\end{tabular}
		}
	\end{minipage}
	\label{table5:2}
    \vspace{-0.2in}
\end{table}

\subsection{Comparisons with State-of-the-Art Methods}
We compare our DTF-Net and DTF-Transformer with various state-of-the-art video backbones on Kinetics-400, Something-Something V1 (SSv1) and V2 (SSv2). All video backbones are grouped into two categories: Convolutional Networks and Vision Transformer.
Here we implement our DTF-Net in two different CNN backbones, i.e., ResNet-50 (R50) and ResNet-101 (R101), and vary the input clip length within the range of \{16, 32, 64\}.
DTF-Transformer is constructed based on the backbone of Swin Transformer (Swin-B) and we fix the input clip length as 64 frames.
We measure the computational cost of each run by GFLOPs $\times$ views (views: the number of clips sampled from the full video at inference).

\begin{table}[!tb]
	\setlength{\belowcaptionskip}{-1pt}
	\caption{Performance comparisons with state-of-the-art video backbones on Something-Something V1 and V2. The input clip length is shown in bracket.}
	\begin{minipage}[!tb]{.5\linewidth}
		\scalebox{0.55}[0.52]{
			\begin{tabular}{{l|l|c|c c|c c}}
				\shline
				\multicolumn{1}{c|}{\multirow{2}{*}{\textbf{Approach}}} & \multirow{2}{*}{\textbf{Backbone}} & {\textbf{GFLOPs}} & \multicolumn{2}{c|}{\textbf{SSv1}}  & \multicolumn{2}{c}{\textbf{SSv2}} \\ \hhline{*{3}{~}*{4}{-}}
				&                                    &           \textbf{$\times$views}            &  \textbf{Top-1}  &  \textbf{Top-5}  &  \textbf{Top-1}  &  \textbf{Top-5} \\ \shline
				\multicolumn{7}{l}{\textbf{Convolutional Networks}} \\ \shline
				\text{NL I3D+GCN \cite{Wang:ECCV18}}                    &  R50       &  606           & 46.1 & 76.8 &  -   &  - \\
				\text{CPNet \cite{Xingyu:CVPR19}}                       &  R34       &  N/A           &  -   &  -   & 57.7 & 84.0 \\
				\text{TSM \cite{JiLin:ICCV19}}                          &  R50       &  98            & 47.2 & 77.1 & 63.4 & 88.5 \\
				\text{TAM \cite{Fan:NIPS19}}                            &  R50       &  48            & 48.4 & 78.8 & 61.7 & 88.1 \\
				\text{GST \cite{Luo:ICCV19}}                            &  R50       &  59            & 48.6 & 77.9 & 62.6 & 87.9 \\
				\text{SmallBig \cite{Li:CVPR20}}                        &  R50       & 105            & 49.3 & 79.5 & 62.3 & 88.5 \\
				\text{CorrNet \cite{Wang:CVPR20}}                       &  R50       & 115$\times$10  & 49.3 &  -   &  -   &  -   \\
				\text{ACTION-Net \cite{WangAct:CVPR21}}                 &  R50       & 69             &  -   &  -   & 64.0 & 89.3 \\
				\text{STM \cite{Jiang:ICCV19}}                          &  R50       & 67$\times$30   & 50.7 & 80.4 & 64.2 & 89.8  \\
				\text{MSNet \cite{Kwon:ECCV20}}                         &  R50       & 67             & 52.1 & 82.3 & 64.7 & 89.4 \\
				\text{TEINet \cite{Liu:AAAI20}}                         &  R50       & 99             & 52.5 &  -   & 65.5 & 89.8   \\
				\text{MG-TEA \cite{Zhi:ICCV21}}                         &  R50       &  N/A           & 53.2 &  -   & 63.8 &  -     \\
				\text{TDN \cite{Wang:CVPR21}}                           &  R50       & 72             & 53.9 & 82.1 & 65.3 & 89.5 \\ \hline
				\text{DTF-Net (16)}							            &  R50	     & 25$\times$3    & 54.2 & 82.3 & 65.5 & 89.6 \\
				\text{DTF-Net (32)}							            &  R50	     & 51$\times$3    & 55.1 & 83.0 & 66.2 & 90.3 \\
				\text{DTF-Net (64)}							            &  R50	     & 111$\times$3   & \textbf{56.2}  & \textbf{83.9}  & \textbf{67.1} & \textbf{90.9}\\ \shline
			\end{tabular}
		}
	\end{minipage}
    \hspace{-0.2in}
	\begin{minipage}[!tb]{.5\linewidth}
		\scalebox{0.63}[0.60]{
			\begin{tabular}{{l|l|c|c c|c c}}
				\shline		
				\multicolumn{1}{c|}{\multirow{2}{*}{\textbf{Approach}}} & \multirow{2}{*}{\textbf{Backbone}} & {\textbf{GFLOPs}} & \multicolumn{2}{c|}{\textbf{SSv1}}  & \multicolumn{2}{c}{\textbf{SSv2}} \\ \hhline{*{3}{~}*{4}{-}}
				&                                    &           \textbf{$\times$views}            &  \textbf{Top-1}  &  \textbf{Top-5}  &  \textbf{Top-1}  &  \textbf{Top-5} \\ \shline		
				\multicolumn{7}{l}{\textbf{Convolutional Networks}} \\ \shline
				\text{GSM \cite{Sudhakaran:CVPR20}}                     &  Inception & 268            & 55.2 & -   & -  &  - \\
				\text{CorrNet \cite{Wang:CVPR20}}                       &  R101      & 224$\times$30  & 53.3 & -   & -  &  - \\
				\text{MG-TEA \cite{Zhi:ICCV21}}                         &  R101      & N/A            & 53.3 & -   & 64.8 & - \\
				\text{TDN \cite{Wang:CVPR21}}                           &  R101      & 132            & 55.3 & 83.3 & 66.9 & 90.9 \\ \hline
				\text{DTF-Net (16)}							            &  R101	     & 38$\times$3    & 55.4 & 83.4 & 67.1 & 91.5 \\
				\text{DTF-Net (32)}							            &  R101	     & 76$\times$3    & 56.4 & 83.8 & 68.2 & 92.3 \\
				\text{DTF-Net (64)}							            &  R101      & 152$\times$3   & \textbf{57.1} & \textbf{84.1}  & \textbf{68.9} & \textbf{92.6}   \\
				\shline
				\multicolumn{7}{l}{\textbf{Vision Transformer}} \\ \shline
				\text{TimeSformer \cite{Bertasius:ICML21}}              &  ViT-B     & 1,703$\times$3 &  -  & -  & 62.5 & - \\
				\text{ViViT \cite{ViViT}}                               &  ViT-L     & 903            &  -  & -  & 65.4 & 89.8 \\
				\text{MViT \cite{Fan:MVIT}}                             &  ViT-B     & 455$\times$3   &  -  & -  & 67.7 & 90.9 \\
				\text{Video-Swin \cite{Liu:V-Swin}}                     &  Swin-B    & 321$\times$3   &  -  & -  & 69.6 & 92.7 \\
				\hline
				\text{DTF-Transformer}                                 &  Swin-B    & 266$\times$3   &  \textbf{57.9} & \textbf{85.7}  & \textbf{70.1}  & \textbf{93.2} \\
				\shline			
			\end{tabular}
		}
	\end{minipage}
	\label{table5:3}
    \vspace{-0.28in}
\end{table}

Table \ref{table5:2} summarizes the performance comparisons on Kinetics-400. In Convolutional Networks group, our DTF-Net achieves better performances than other baselines.
In particular, DTF-Net (32) in R50 obtains 78.9\% top-1 accuracy, surpassing the best competitor TDN by 1.4\% and relatively reducing 30\% computation cost in GFLOPs.
Note that although TDN emphasizes the long-term temporal structure by cross-segment feature enhancement, its temporal receptive field is still restricted by the traditional block design.
In contrast, our DTF is benefited from the mechanism of dynamic temporal modeling in frequency domain with enlarged temporal receptive field.
DTF-Net (64) further improves the top-1 accuracy from 78.9\% to 80.9\% by exploiting more frames in each clip.
When inserting DTF block into the advanced 2D Vision Transformer (Swin Transformer), DTF-Transformer achieves the best performance (83.5\%) in top-1 accuracy.
In comparison to the superior 3D Vision Transformer (Video-Swin), DTF-Transformer leads to 0.8\% performance gain and with less computation cost.
This basically verifies the better temporal modeling of spatial-aware feature spectrum filtering than the self-attention along temporal dimension.

Table \ref{table5:3} shows the performances on SSv1 and SSv2 datasets, where the common one-clip and 3-crops settings \cite{Bertasius:ICML21,Fan:MVIT,Liu:V-Swin} are adopted for evaluation.
Similar performance trends are observed on the two datasets, and DTF-Net (64) in R101 backbones outperforms the best competitor TDN by 1.8\% and 2.0\% top-1 accuracy on SSv1 and SSv2, respectively.
By further plugging DTF block into the Swin-B backbone, DTF-Transformer obtains the best performances on both datasets, confirming the superiority of our DTF block in video modeling.

\subsection{Visualization Analysis of Dynamic Temporal Filter Block}
\begin{figure}[!tb]
	\centering\includegraphics[width=0.98\textwidth]{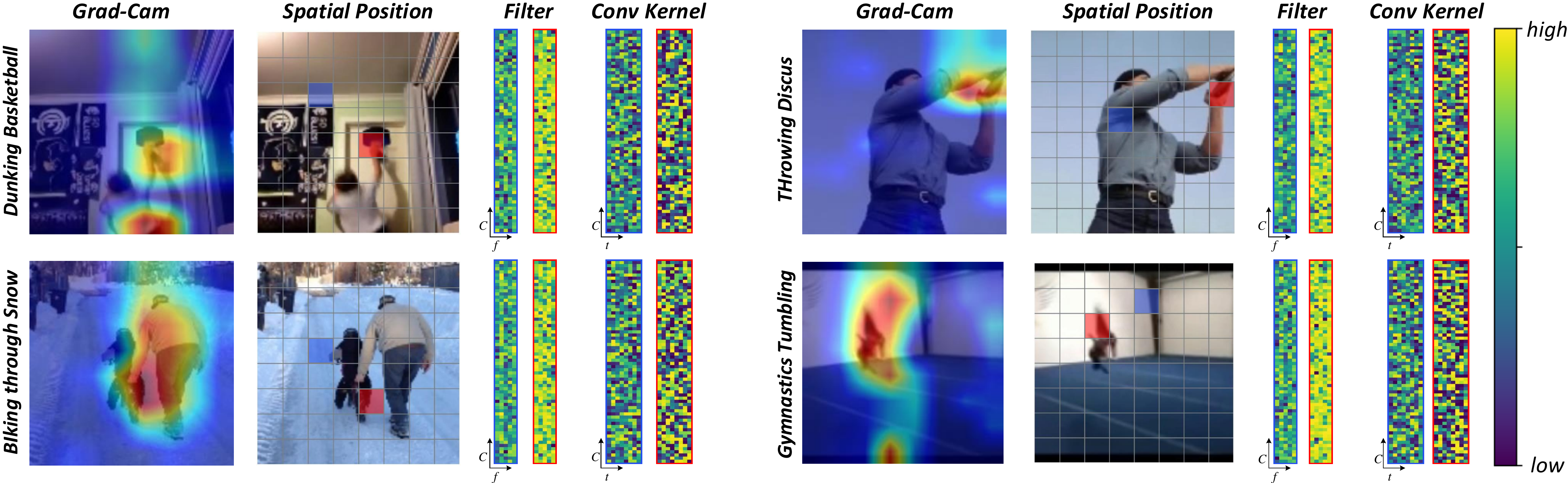}
    \vspace{-0.1in}
	\caption{Visualization of the Grad-CAM \cite{Grad-CAM}, two selected spatial positions, the learnt frequency filter and the corresponding 1D convolution kernel of DTF-Net in each position for four Kinetics-400 videos. We select two positions of each video based on Grad-CAM, where the blue and red box represents the position with small and large movements, respectively. The 1D convolution kernel is obtained by applying IFFT over the learnt  filter. The visualization of frequency filter or 1D kernel is marked with box in the same color with the corresponding position.}
	\label{fig1:7}
\vspace{-0.2in}
\end{figure}

To better qualitatively analyze the temporal modeling of DTF block, we further visualize the class activation map with Grad-CAM \cite{Grad-CAM}, two selected spatial positions, the learnt frequency filter and the corresponding 1D convolution kernel for each selected spatial position in DTF-Net (backbone: R50) in Figure \ref{fig1:7}.
Note that Grad-CAM naturally reflects the meaningful motion cues that benefit action recognition, where the region with larger class activation response commonly refers to spatial position with larger movements tailored to the target action. Therefore we sample two spatial positions according to the class activation of Grad-CAM: one with large movement (in red box) and the other with small movement (in blue box). Next, for each spatial position, we visualize its frequency filter in $res_5$ stage of DTF-Net (resolution: $8 \times 8$), and the corresponding 1D convolution kernel is calculated by IFFT over frequency filter. Specifically, for each video, the learnt frequency filter/1D convolution kernel of spatial position with large movement is clearly more active than that of location with small movement. The results validate that DTF block effectively captures differences of spatial contexts at varied locations, and learns a specialized frequency filter for each spatial location, leading to a dynamic spatial-aware temporal modeling.

\section{Conclusions}
In this work, we present a new Dynamic Temporal Filter (DTF) block that formulates dynamic temporal modeling in the frequency domain with an enlarged temporal receptive field. Particularly, DTF mechanism first takes all features across time in the same spatial location as temporal feature, and further learns specialized frequency filter based on the temporal feature. Next, the primary temporal feature is transformed into frequency feature spectrum via FFT, which are modulated by the learnt frequency filter. The modulated frequency spectrum is finally transformed back to temporal domain via IFFT. Going beyond DTF mechanism, DTF block additionally employs frame-wise aggregation module to not only contextualize temporal feature but also enable more effective learning of frequency filter.
By plugging DTF block into ResNet and Swin-Transformer, we construct two new video backbones, i.e., DTF-Net and DTF-Transformer.
Experiments conducted on three action recognition datasets demonstrate the superiority of both DTF-Net and DTF-Transformer.

\textbf{Acknowledgments.} This work was supported by the National Key R\&D Program of China under Grant No. 2020AAA0108600.

%
%
\bibliographystyle{splncs04}
\bibliography{egbib}
\end{document}